\documentclass[runningheads]{llncs}
\usepackage{graphicx}
\usepackage{colortbl,booktabs}
\usepackage{amsmath}
\usepackage{arydshln}
\usepackage{multirow}
\usepackage{soul}
\usepackage{tablefootnote}
\usepackage{lipsum} 
\usepackage{pifont}
\newcommand{\xmark}{\ding{55}}
\newcommand{\cmark}{\ding{51}}
\renewcommand\hl[1]{#1} 
\begin{document}
\title{Adversarially Robust Deepfake Detection via Adversarial Feature Similarity Learning}

\author{Sarwar Khan\inst{1,2,3}
\and
Jun-Cheng Chen\inst{1,2}
\and
Wen-Hung Liao\inst{2,3}
\and
Chu-Song Chen\inst{4}
}

\authorrunning{S. Khan et al.}
\institute{Research Center for Information Technology Innovation, Academia Sinica\and
Social Networks Human-Centered Computing, TIGP, Academia Sinica\\
\and
Computer Science, National Chengchi University \\
\and
Computer Science and Information Engineering, National Taiwan University\\
say2sarwar@gmail.com, pullpull@citi.sinica.edu.tw, whliao@cs.nccu.edu.tw, chusong@csie.ntu.edu.tw
}

\titlerunning{AFSL}

\maketitle
\begin{abstract}
Deepfake technology has raised concerns about the authenticity of digital content, necessitating the development of effective detection methods. 
However, the widespread availability of deepfakes has given rise to a new challenge in the form of adversarial attacks.
Adversaries can manipulate deepfake videos with small, imperceptible perturbations that can deceive the detection models into producing incorrect outputs.
To tackle this critical issue, we introduce Adversarial Feature Similarity Learning (AFSL), which integrates three fundamental deep feature learning paradigms.
By optimizing the similarity between samples and weight vectors, our approach aims to distinguish between real and fake instances. 
Additionally, we aim to maximize the similarity between both adversarially perturbed examples and unperturbed examples, regardless of their real or fake nature. 
Moreover, we introduce a regularization technique that maximizes the dissimilarity between real and fake samples, ensuring a clear separation between these two categories.
With extensive experiments on popular deepfake datasets, including FaceForensics++, FaceShifter, and DeeperForensics, the proposed method outperforms other standard adversarial training-based defense methods significantly. This further demonstrates the effectiveness of our approach to protecting deepfake detectors from adversarial attacks.

\end{abstract}

\section{Introduction}
Deepfakes are synthetic videos in which a person's face is altered to resemble a different individual, resulting in the production of highly realistic footage depicting events that never actually took place~\cite{Deepfakes}.
Deepfake technology has captivated and alarmed society by offering the ability to create remarkably convincing and misleading media, which in turn threatens the authenticity of digital content. 
In response to these concerns, researchers have diligently worked to develop effective deepfake detection methods~\cite{afchar2018mesonet,Dong_2023_CVPR,haliassos2022RealForensics,haliassos2021lips,rossler2019faceforensics++,shahzad2022lip,zheng2021exploring}.\\
\indent Deepfake detectors have shown promising performance under normal conditions, accurately identifying manipulated videos. 
However, a new challenge has emerged in the form of adversarial attacks, where small and imperceptible perturbations can deceive the detection models into producing incorrect outputs~\cite{gandhi2020adversarial,guan2022delving,hou2023evading,hussain2021adversarial,Liu_2023_WACV,neekhara2021adversarial}. 
An adversarial example is a manipulated input intentionally designed to deceive a classification model~\cite{madry2017towards}.
Adversarial deepfakes~\cite{hussain2021adversarial} leverage the pre-softmax layer to compute the loss and iteratively calculate gradients, allowing for the creation of adversarial fakes that can successfully evade detection.
Statistical consistency attack (StatAttack) robust\cite{hou2023evading} looks into statistical consistency between real and fake and uses degradation techniques to create transferable deepfake adversarial attacks.
This poses a significant threat to the reliability and effectiveness of deepfake detection systems, as it undermines their ability to distinguish between genuine and manipulated content.\\
\indent \hl{Adversarial training\mbox{~\cite{madry2017towards}} tackles adversarial attacks through a \textit{min-max} optimization but often at the cost of reduced performance on normal inputs. To address this, TRADES\mbox{~\cite{zhang2019theoretically}} is a surrogate loss for adversarial training utilizing cross-entropy loss supervising and the distance loss between the features of clean and adversarial examples as the regularization. However, training deepfake detectors for robustness remains challenging due to the inclusion of fake images. Our study is motivated by the recognition that adversaries can exploit misclassification, aiming to develop effective strategies for adversarially robust deepfake detection.}\\
\indent In pursuit of this objective, we develop an Adversarial Feature Similarity Learning (AFSL) objective function that optimizes similarity across three fundamental paradigms of deep feature learning.
First, we optimize the similarity between samples and weight vectors, specifically focusing on differentiating between real and fake instances. 
Secondly, our objective is to maximize the similarity between samples, considering both adversarially perturbed examples and unperturbed examples, where the perturbed instances can be either real or fake. 
Finally, we introduce a regularization approach that aims to maximize the dissimilarity between real and fake samples, ensuring a clear separation and distinct representation of these two categories. 
This comprehensive approach enables effective deepfake detection by enhancing discrimination between real and fake content and mitigating the impact of adversarial perturbations.
We conduct extensive experiments on the FaceForensics++~\cite{rossler2019faceforensics++}, FaceShifter~\cite{li2020advancing}, and DeeperForensics~\cite{jiang2020deeperforensics} datasets, evaluating the performance of our proposed method. Impressively, our method 
outperforms widely used adversarial training-based defense methods by a significant margin. This demonstrates the effectiveness of our approach to help deepfake detectors fight against various adversarial attacks.
\section{Related work}
\subsection{Deepfake Creation and Detection}
The development of Generative Adversarial Networks (GANs) and their diverse variants has yielded remarkable outcomes in image generation and manipulation, consequently facilitating the emergence of deepfake technology. 
By leveraging GANs, deepfake has enabled the creation of fabricated images or videos across various categories.
The current deepfake generation techniques include various approaches, such as complete face synthesis~\cite{karras2019style}, face identity swap~\cite{gao2021information}, and face manipulation~\cite{gao2021high}.
The utilization of these generation methods by malicious applications can greatly jeopardize public information security. Nonetheless, it is crucial to recognize that the misuse of deepfake technology raises additional concerns regarding security and privacy, extending to sensitive areas such as politics, religion, and pornography~\cite{spivak2018deepfakes,yadlin2021whose}.
\hl{Meanwhile study conducted in Thailand \mbox{\cite{songja2023deepfake}} explores Thai perspectives on deepfake AI images, emphasizing both creative interests and potential risks. It suggests the Thai government take a proactive role in regulating and raising awareness to harness the technology's creative potential while addressing concerns related to data protection and image copyright.}\par
To mitigate the potential misuse of deepfake technologies, various Deep Neural Network (DNN) methods have been proposed for detecting deepfake inputs. 
Deepfake detection primarily involves the binary classification of distinguishing between fake and real inputs. 
In the realm of deepfake detection methods, some methods focus on extracting spatial information~\cite{Dong_2023_CVPR,guan2022delving,haliassos2022RealForensics,haliassos2021lips,Liu_2023_WACV,zheng2021exploring}, whereas others delve into analyzing the differences in frequency information~\cite{frank2020leveraging} between fake and real inputs.
LipForensics~\cite{haliassos2021lips} employs a lips extraction technique from facial images and leverages a combination of a pretrained feature extractor and a temporal convolutional network to train an effective deepfake detector.
FTCN~\cite{zheng2021exploring} proposed exceptional generalization across various manipulation scenarios by enforcing a uniform spatial convolutional kernel size of one.
RealForensics~\cite{haliassos2022RealForensics} aims to enhance forgery detection performance and improve generalization across different datasets by leveraging real talking faces through self-supervision using spatiotemporal features.
These methods achieve remarkable detection results within their respective experimental configurations by harnessing the formidable feature extraction capabilities of DNNs.
\subsection{Adversarial Examples}
Adversarial examples exploit the vulnerability of deep learning models by intentionally designing inputs that cause the models to make mistakes or misclassify data~\cite{carlini2017adversarial}.
Gradient-based adversarial attacks are extremely effective against deep learning models in image~\cite{carlini2017adversarial,Li_2023_CVPR,Liang_2023_CVPR,madry2017towards,khan2017text}, video~\cite{mumcu2022adversarial,wang2021understanding}, and audio~\cite{chen2022towards,neekhara2019universal,qin2019imperceptible} domain. 
FGSM~\cite{goodfellow2014explaining}, PGD~\cite{madry2017towards}, CW~\cite{carlini2017adversarial}, and StatAttack~\cite{hou2023evading} represent potent adversarial attack techniques that employ distinct optimization strategies to generate perturbations capable of deceiving classification models.
Similar to other deep learning models, deepfake detectors are vulnerable to adversarial attacks, making them a significant threat within the realm of deepfakes.\par

Adversarial deepfakes~\cite{hussain2021adversarial} present a robust white-box attack (RWB) setting, where the attacker possesses full access to the model, as well as a robust black-box (RBB) attack setting, where the attacker lacks access to the model. 
These settings are achieved by utilizing the pre-softmax layer and employing diverse transformations.
Likewise, Neekhara et al.~\cite{neekhara2021adversarial} investigate the transferability of adversarial attacks in forgery detectors and propose a universal attack approach, demonstrating the effectiveness of adversarial examples across different models and architectures.
In addition, Statistical Consistency Attack (StatAttack)~\cite{hou2023evading} introduces a transferable approach that leverages adversarial statistical consistency through the minimization of a distribution-aware loss, enabling it to circumvent deepfake detectors effectively.\par
A multitude of defense strategies have been proposed to combat adversarial examples in both image and video domains, adversarial training has exhibited commendable performance in mitigating the impact of such attacks~\cite{madry2017towards,zhang2019theoretically,lo2021defending,ACL2020,yang2021defending}.
Although adversarial training has shown effectiveness against various forms of adversarial attacks, it often comes at the cost of reduced performance on unperturbed data. 
To address this, Deep image prior (DIP)~\cite{gandhi2020adversarial} utilizes a GAN network to remove perturbations from the input, making the model robust against adversarial attacks, and also incorporates regularization techniques as a defense strategy in deepfake detection.
\hl{To handle another growing concern of deepfake technology exacerbated by the use of face masks during the pandemic, Alnaim et al. \mbox{\cite{alnaim2023dffmd}} propose 
a novel deepfake face mask dataset and detection model with identifying face-mask-related deepfakes.}
To enhance the robustness of audio deepfake detection, audio-based models employ adversarial training and adaptive training techniques~\cite{kawa2023defense}, focusing solely on the audio modality.
Previous research has demonstrated the effectiveness of defending against various attack methods by utilizing projected gradient descent (PGD) as a defense mechanism, validating the proposal by Madry et al.,~\cite{madry2017towards}.
This further supports the notion that robust defenses, specifically tailored to counter PGD attacks, can provide efficient protection against different first-order attack methods.\par 
\section{Adversarial Feature Similarity Learning}

\textbf{Notation:} We consider $f_\theta(\cdot)$ as the feature encoder from the deepfake detector and $\theta$ representing its learnable model parameters.
The variable $x \in \mathcal{X}$ denotes an input frame or a video clip from a video depending on if $f_\theta(\cdot)$ is a frame-based or a video-based deepfake detector method, where $x$ can be either real or fake samples. 
In addition, $x_{\text{adv}}$ represents the adversarial example generated using $x \in \{x^{real},x^{fake}\}$ where $x^{real} \in \mathcal{X}$ and  $x^{fake} \in \mathcal{X}$ represent real and fake samples, which can be either an individual frame or a video clip depending on whether $f_\theta(\cdot)$ is a frame-based or video-based detector. 
$y \in {\{0,1\}}$ denotes the label, where 0 indicates a fake class and 1 real class.
Adversarial Feature Similarity Learning
\subsection{Overview}
\begin{figure}[t]
  \centering
  \includegraphics[scale=0.61]{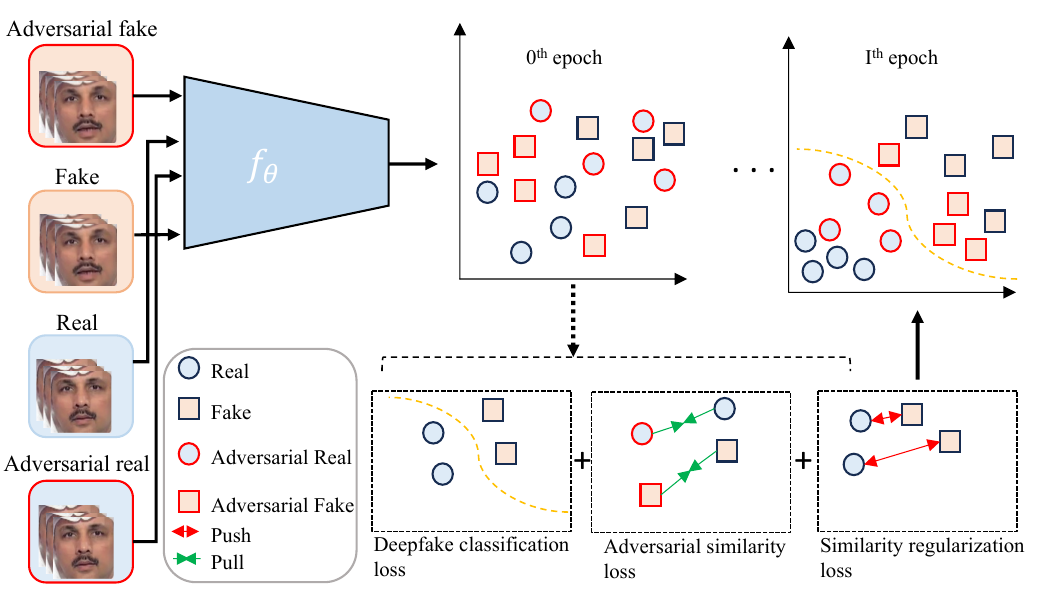}
  \caption{Framework for adversarial feature similarity learning. First, we select a pair of real and deepfake samples and create adversarial perturbation for the corresponding inputs. Then, we generate the features of real, fake, and their adversarial samples. Finally, through the proposed loss function, the model can learn a better representation to separate real samples from fake ones along with their adversarial counterparts, where the backbone $f_{\theta}$ is from the deepfake detector.  }
  \label{fig:main}
\end{figure}

Our objective is to develop an adversarially robust deepfake detector that effectively mitigates the impact of adversarial attacks while preserving the performance on unperturbed data.
We address this problem by discerning the features of real and fake samples and their corresponding adversarial counterparts. 
We hypothesize that an adversarial attack will shift the features in the opposite direction, irrespective of whether the input is real or fake.
However, deepfake detectors trained solely with adversarial training will not effectively learn the desirable features to distinguish between real and fake samples in the presence of adversarial attacks.
To overcome this limitation, we proposed a novel loss function to effectively separate the two classes i.e. (Real and Fake) under most conditions. 
Figure \ref{fig:main} provides a comprehensive illustration of our framework.
We adopt a three-step approach.
Separating unperturbed (Real and Fake) in Section \ref{LABCE}, using adversarial similarity loss to make the detector robust in Section \ref{GAP}, and finally similarity regularization loss to preserve the unperturbed performance in Section \ref{reg}.
Additionally, we formulate the final loss function in Section \ref{Formulation}.
\subsection{Deepfake Classification Loss}
\label{LABCE}
The deepfake classification loss is realized using a supervised loss that utilizes a logit-adjusted variant of binary cross-entropy (LBCE)~\cite{menon2021logitadjustment}, denoted as $\mathcal{L}_{LBCE} \left(f_\theta(x),y \right)$ to tackle the potential issues of class imbalance.
The supervised deepfake classification loss function aims to maximize the dissimilarity between real and fake samples, while simultaneously addressing the class imbalance in the dataset. 
This approach helps to refine the model's discrimination abilities, enabling it to better differentiate between real and fake videos.
The deepfake classification loss, $\mathcal{L}_{\text{dcl}}$, is denoted as follows:
\begin{equation}
    \mathcal{L}_{\text{dcl}}= \mathcal{L}_{LBCE} \left(f_\theta(x),y \right),
\end{equation}
\subsection{Adversarial Similarity Loss}
\label{GAP}
Let \(f_\theta(x)\) and \(f_\theta(x_{\text{adv}})\) respectively represent the mapping from an input sample \(x \in \{x^{\text{real}}, x^{\text{fake}}\}\) and from an adversarial input sample \(x_{\text{adv}} \in \{x^{\text{real}}_{\text{adv}}, x^{\text{fake}}_{\text{adv}}\}\) to their corresponding embedding spaces.
To perform adversarial training, we generate adversarial examples $x_{\text{adv}}$ from $x\in \{x^{\text{real}}, x^{\text{fake}} \}$ by employing the PGD adversarial attack method. 
The objective is to maximize the cosine similarity $sim\left(f_\theta(x),f_\theta(x_{\text{adv}})\right))$ between $x$ and $x_{\text{adv}}$ to bring them closer as they represent the same class and to avoid adversarial examples from being misclassified to the other class (i.e., real to fake and fake to real).
To achieve this objective while minimizing the final loss, we introduce adversarial similarity loss as follows
\begin{equation}
    \mathcal{L}_{asl}= (1 -sim\left(f_\theta(x),f_\theta(x_{\text{adv}})\right)),
\end{equation}
\noindent where $sim(\cdot,\cdot)$ indicates the cosine similarity metric, and $\mathcal{L}_{asl}$ is the adversarial similarity loss.
We aim to minimize the dissimilarity using adversarial similarity loss, thereby maximizing the similarity.
\subsection{Similarity Regularization Loss}
\label{reg}
To further enhance the performance, we enforce additional regularization to minimize the similarity between the paired real and fake samples from the corresponding real and deepfake samples. 
The similarity regularization loss minimizes similarity, which is computed through cosine similarity using the unperturbed samples from real and deepfake inputs.
This process effectively creates separation between the two classes, ensuring that the detector's unperturbed performance remains intact. 
By employing this approach, we not only improve the detector's robustness but also preserve its unperturbed performance.
Similarity regularized loss is calculated as follows:
\begin{equation}
    \mathcal{L}_{\text{srl}}= sim\left(f_\theta(x^{\text{real}}),f_\theta(x^{\text{fake}})\right),
\end{equation}
\noindent where $\mathcal{L}_{\text{srl}}$ is the similarity regularization loss, $ \mathcal{L}$ is cosine similarity, and respectively $x^{\text{real}},x^{\text{fake}} \in \mathcal{X}$ are real and fake pair input. 
 
\subsection{Final Loss Function}
\label{Formulation}

In this section, we formulate the final loss function for minimization based on the three previously discussed components.
The objective function is presented in Equation \ref{eq:main}.
\begin{equation} \label{eq:main}
\mathcal{L}_{\text{afsl}}=\mathcal{L}_{\text{dcl}}+\beta_1\mathcal{L}_{\text{asl}}+ \beta_2\mathcal{L}_{\text{srl}},
\end{equation}
where $\mathcal{L}_{\text{dcl}}$ denotes the deepfake classification loss, while $\mathcal{L}_{\text{asl}}$ and $\mathcal{L}_{\text{srl}}$ correspond to the adversarial similarity loss and similarity regularization loss respectively.  
To control the influence of the regularization terms, we set the scaling factors $\beta_1$ and $\beta_2$ to 1 and 0.1, respectively.
By incorporating these components into the loss function $\mathcal{L}_{\text{afsl}}$, we aim to enhance the robustness of the detectors while preserving their unperturbed performance.\par
Unlike TRADES, our approach utilizes similarity regularization, capturing finer differences between real and fake videos and thereby enabling the extraction of more accurate features.

\setlength{\tabcolsep}{6.3pt}
\begin{table}[t]
    \centering
    \caption{Video level AUC (\%) for deepfake detectors when testing on each deepfake type of FF++ after training on the remaining three types. ``No Attack'' denotes that an adversarial attack is not applied while ``PGD10'' denotes that the Projected Gradient descent (PGD) attack is applied to the input. The types of deepfakes are DeepFakes (DF), FaceSwap (FS), Face2Face (F2F), and NeuralTextures (NT). \textbf{All the numbers of the baseline methods shown in this Table are reproduced based on the default settings of the officially released implementations, and the performance discrepancies from their papers may be due to the released versions or hyperparameters being different from the ones used in the experiments of the papers.}}
    \begin{tabular}{l|cccc|cccc}
         \toprule
         \multirow{2}{*}{Methods}&\multicolumn{4}{c}{\textbf{No Attack}}&\multicolumn{4}{|c}{\textbf{PGD10}}\\
         \cmidrule{2-4}\cmidrule{5-9}
         &DF&FS&F2F&NT&DF&FS&F2F&NT  \\
         \midrule
         RealForensics~\cite{haliassos2022RealForensics}&91.5&89.6&92.3&92.6 &0.7&0.8&1.2&1.6\\
         LipForensics~\cite{haliassos2021lips}&90.8&87.1&92.0&91.8 &2.8&2.4&1.0&2.6\\
         FTCN~\cite{zheng2021exploring} &91.6&90.0&91.6&92.8&1.0&1.2&1.7&2.4\\
         \hdashline
         Patch-based~\cite{chai2020makes}&85.7&57.4&84.6&80.3&3.8&2.6&1.8&4.2\\
         Xception~\cite{rossler2019faceforensics++}&83.1&50.6&81.5&76.9&1.0&4.7&0.8&0.2\\
         
         \bottomrule
    \end{tabular}
    
    \label{tab:all_det}
\end{table}
\section{Experimental Description}
\subsection{Implementation Details}
The faces are extracted through the utilization of face detection and alignment techniques and video clips comprising 25 frames.
For the training process, the video clips are randomly cropped to dimensions $140\times140$ and subsequently resized to $112\times112$.
Horizontal flipping and grayscale transformation with a probability of 0.5 are applied along with random masking. 
\hl{For sequence-based deepfake detection, we employ the Channel-Separated Convolutional Network (CSN) \mbox{\cite{tran2019video}}.
We refer interested readers to \mbox{\cite{haliassos2022RealForensics,tran2019video}} for more details about CSN.}
The optimization process utilizes the Adam optimizer with a learning rate of $3 \times 10^{-4}$.
The model is trained for 150 epochs.
We used RealForensics self-supervised pretrained on Lip Reading in the Wild (LRW) dataset \footnote{\label{realforensics}Pretrained model: https://github.com/ahaliassos/RealForensics}.
This pretrained model serves as the starting point and provides the initial push to effectively capture the relevant features, thereby enhancing the model's generalization capabilities.

While for frame-based detection, we utilize XceptionNet~\cite{chollet2017xception} and MesoNet~\cite{afchar2018mesonet} \footnote{\label{xceptionnet}Pretrained model: https://github.com/paarthneekhara/AdversarialDeepFakes}.
We optimize the frame-based model using a Stochastic gradient descent (SGD) optimizer with a learning rate of $2 \times10^{-3}$ and the model is trained for 150 epochs with a batch size of 16.
\hl{For further details about the frame-based model, interested readers are directed to \mbox{\cite{afchar2018mesonet,advfake}}.}
Normalization ($L_2$-norm) is applied to all features in both sequence-based and frame-based detectors. 

\noindent\textbf{Datasets:} FaceForensics++ (FF++), is comprised of 1,000 authentic videos and 4,000 deepfake videos. 
Unless specified otherwise, the mildly compressed version of the dataset (c23) was utilized.
Other datasets used in the experiments are FaceShifter~\cite{li2020advancing} and DeeperForensics~\cite{jiang2020deeperforensics}, featuring different face-swapping techniques applied to FF++ real videos.

\noindent\textbf{Evaluation metrics:} We utilize accuracy and area under the receiver operating characteristic curve (AUC) metrics. 
For video-level assessment, we uniformly sample non-overlapping clips from a single video and average their predictions.
\setlength{\tabcolsep}{6.pt}
\begin{table}[t]
    \centering
    \caption{AUC (\%) scores for video-level detection on the FF++ dataset, containing four deepfake methods. Models train on three methods and test on the remaining method. We employ PGD5 for adversarial training and PGD10 for testing purposes. $L_\infty$ is allowed distortion for adversarial attacks. $L_\infty =0$Adversarial Feature Similarity Learning means no adversarial attack is applied.}
    \begin{tabular}{l|c|ccccc}
    \toprule
    Method&$L_\infty$&DF&FS&F2F&NT&Avg\\
    \midrule
    RealForensics\cite{haliassos2022RealForensics}&0&91.5&89.6&92.3&92.6&91.5\\
    RealForensics &8/255&0.7&0.8&1.2&1.6&1.0\\
    RealForensics + AT~\cite{madry2017towards} &8/255&76.3&74.1&73.7&70.1&73.5\\
    RealForensics + TRADES~\cite{zhang2019theoretically}&8/255&78.2&75.4&80.7&72.5&76.7\\
    \midrule
    RealForensics + AFSL (Ours)&0&89.4&87.6&90.4&91.7&89.7\\
    AFSL (Ours)&8/255&79.0&77.2&82.6&75.6&78.6\\
    RealForensics + AFSL (Ours) &8/255 &81.5&79.7&84.1&78.1&80.8\\
    \bottomrule

    \end{tabular}
    
    \label{tab:one_out}
\end{table}

\subsection{Victim Models: Deepfake Detectors}
In our work, we assess the vulnerability of top-performing deepfake detectors to adversarial attacks. 
We employ video-based detectors, namely RealForensics~\cite{haliassos2022RealForensics}, LipForensics~\cite{haliassos2021lips}, and FTCN~\cite{zheng2021exploring}. 
In addition, we incorporate frame-by-frame based detectors, such as Patch-based~\cite{chai2020makes} and Xception~\cite{rossler2019faceforensics++}. 
All the models are tested on each of the four methods using FF++ after training on the remaining three. 
Table \ref{tab:all_det} presents the AUC score of all detectors under two conditions: ``No Attack,'' where no adversarial attack is applied to the input, and ``PGD10,'' where an adversarial attack is applied to the input video.

As observed from the results, all the deepfake detectors demonstrate vulnerability to adversarial attacks. 
Our proposed loss function offers the advantage of seamless integration with most deepfake detectors.
For the evaluation of our method, we select RealForensics from sequence-based detectors, along with XceptionNet and MesoNet from frame-by-frame detectors.
As we cannot evaluate every detector, we choose the top-performing detector from each category.\par
We employ pretrained weights from RealForensics$^{\ref{realforensics}}$ and fine-tune LipForensics, FTCN\footnote{https://github.com/yinglinzheng/FTCN}, Patch-based\footnote{https://github.com/chail/patch-forensics}, and Xception. 
We follow the exact instructions for pre-processing provided in their official code to replicate the results. 
While we do observe a decline in performance compared to the reported results, this could potentially be attributed to the absence of supplementary data. 
We solely report the reproduced results as we aim to improve model robustness against adversarial attacks.

\setlength{\tabcolsep}{3pt}
\begin{table*}[t]
    \centering
    \caption{Average video-level AUC (\%) for adversarial attacks is computed by training the model on three methods and testing it on a fourth method. The reported values represent the average scores across the entire test dataset, encompassing all four methods, under both white-box and black-box adversarial attacks.}
    \resizebox{\textwidth}{!}{%
    \begin{tabular}{lccccccc}
    \toprule
         Methods&No Attack&PGD10&RWA~\cite{hussain2021adversarial}&CW2&SA\cite{hou2023evading}&Ul~\cite{neekhara2021adversarial}&RBB~\cite{hussain2021adversarial}\\
         \midrule
         RealForensics~\cite{haliassos2022RealForensics} & \textbf{91.5}&1.1&1.4&0.0&0.0&0.0&36.8\\
         RealForensics + AT~\cite{madry2017towards}&78.4&73.5 &79.5 &78.3&63.6&66.9&80.5\\ 
         RealForensics + TRADES~\cite{zhang2019theoretically}&84.0& 76.7&76.1&78.1&67.2&69.3&82.4\\
         \hline
         AFSL (Ours) &87.3&78.3&79.1& 79.7&68.5&68.7&86.1\\
         RealForensics +AFSL (Ours)&89.8& \textbf{80.7}&\textbf{81.3}&\textbf{82.8}&\textbf{73.9}&\textbf{74.7}& \textbf{87.5}\\

         \bottomrule
    \end{tabular}}
    
    \label{tab:leave_one_out_avg}
\end{table*}
\begin{table*}[t]
    \centering
    \caption{Video level AUC(\%) for unseen datasets: DeeperForensics and FaceShifter under different adversarial attacks.} 
    \resizebox{\textwidth}{!}{%
    \begin{tabular}{lccccccc}
    \toprule
         Methods&No Attack&PGD10&RWA~\cite{hussain2021adversarial}&CW2&SA~\cite{hou2023evading}&Ul~\cite{neekhara2021adversarial}&RBB~\cite{hussain2021adversarial}\\
         \midrule
         \multicolumn{8}{c}{\textbf{DeeperForensics}}\\
         \midrule
         RealForensics~\cite{haliassos2022RealForensics} & \textbf{93.6}&1.0&0.0&0.0&0.0&0.0&42.2\\
         RealForensics + AT~\cite{madry2017towards}&84.5&78.2  &76.4&75.0&65.4& 69.7&81.3\\
         RealForensics + TRADES~\cite{zhang2019theoretically}&88.1& 78.0&79.3&77.9&69.7&\textbf{85.6}&85.8\\
         \hline
         AFSL (Ours)&90.2&80.3&79.9&80.4&69.4&83.5&87.3\\
         RealForensics +AFSL (Ours)&92.9& \textbf{83.6}& \textbf{81.2}&\textbf{83.5}&\textbf{72.8} &85.6&\textbf{89.1}\\
         \midrule
         \multicolumn{8}{c}{\textbf{FaceShifter}}\\
         \midrule
         RealForensics~\cite{haliassos2022RealForensics} &\textbf{91.7}&1.0&1.0&1.0&0.0&1.0&37.7\\
   
         RealForensics +  AT~\cite{madry2017towards}& 83.6&76.2&74.0&75.8&60.7&73.1&79.9\\
         RealForensics + TRADES~\cite{zhang2019theoretically} &87.1&79.3&77.9&78.2&\textbf{67.3}&72.8&84.6\\
         \hline
         AFSL (Ours)&88.2&79.6&78.1&81.4&65.1&74.6&84.3\\
         RealForensics +AFSL (Ours)&89.4&\textbf{81.7} &\textbf{80.7}&\textbf{84.6}&67.1&\textbf{78.1}& \textbf{86.5}\\

         \bottomrule
    \end{tabular}}
    
    \label{tab:cross_data_res}
\end{table*}
\setlength{\tabcolsep}{3pt}
\subsection{Robust Cross-Manipulation Generalization}
Most deepfake detectors typically conduct generalization experiments to assess their performance. These experiments involve training the detectors on three methods and testing them on the remaining techniques using the FF++ dataset~\cite{haliassos2022RealForensics,shahzad2022lip}. 
In this study, we follow the same protocol and introduce adversarial attacks during both the training and testing stages, utilizing a sequence-based model.
To make the comparison fair, we utilized self-supervised pretrained weights of RealForensics for all methods using the CSN model.
The results in Table \ref{tab:one_out} demonstrate that our proposed method achieves adversarially robust generalization to unseen adversarial deepfakes. 
While RealForensics achieves the best result with unperturbed input, it fails to withstand adversarial attacks.
On the other hand, our proposed method performs well under both adversarially perturbed and unperturbed data compared with adversarial training (AT) and TRADES.
\hl{AT is the baseline while TRADES is the state-of-the-art method 
in terms of
both clean and robust performance. }
Table \ref{tab:leave_one_out_avg} presents the average AUC score on FF++ using white-box and black-box attacks.
\setlength{\tabcolsep}{7pt}
\begin{table*}[t]
  \caption{Frame-level Accuracy (\%) of deepfake detector on FF++ dataset under different adversarial attacks using XceptionNet and MesoNet models. }
  \label{tab:2dframes}
  \begin{tabular}{lccccc}
    \toprule
    \multicolumn{6}{c}{\textbf{XceptionNet}}\\
    \cmidrule{1-6}
    Methods&PGD10&CW2&RWA~\cite{hussain2021adversarial}&SA~\cite{hou2023evading}&RBB~\cite{hussain2021adversarial}\\
    \midrule
    Adversarial Deepfakes~\cite{hussain2021adversarial}& 7.3&8.7&1.9&0.0&48.3\\
    Vanilla AT~\cite{madry2017towards}&76.0&72.7& 73.1&58.9&80.1\\
    TRADES~\cite{zhang2019theoretically}&80.8&79.4&77.2&76.8&84.7\\
    \hline
    AFSL (Ours)&\textbf{81.7}&\textbf{82.9}&\textbf{80.6}&\textbf{77.5}&\textbf{85.7}\\ 
    \toprule
    \multicolumn{6}{c}{\textbf{MesoNet}}\\
    \midrule
    Adversarial Deepfakes~\cite{hussain2021adversarial}&6.2&7.3&0.0&0.0&44.6\\
    Vanilla AT~\cite{madry2017towards}&74.3&70.7&71.9&55.6&67.4\\
    TRADES~\cite{zhang2019theoretically}&78.2&\textbf{77.1}&72.9&\textbf{73.7}&79.3\\
    \hline
    AFSL (Ours)&\textbf{79.5}&76.9&\textbf{74.1}&72.6&\textbf{81.3}\\ 

  \bottomrule
\end{tabular}
\end{table*}
RBB is a black-box attack generated using ResNet3D~\cite{tran2018closer}, while PGD, CW, StatAttack, Universal, and RWA are white-box attacks.
Our proposed method AFSL outperforms all previous defense techniques across all types of adversarial attacks. 
\par
We also evaluate the robust generalization across datasets by training the model on FF++ using all manipulation methods and testing it on two datasets, DeeperForensics and Faceshifter. Table \ref{tab:cross_data_res} presents the AUC results for both datasets.
We compare the proposed method with state-of-the-art under stronger white-box attacks and black-box attacks. 
The robust AUC confirms that the proposed method performs well compared to the baseline method and other defenses when exposed to various types of adversarial attacks.
\subsection{Evaluation on Frame-based Detectors}
To showcase the effectiveness of our proposed approach, we incorporated two frame-by-frame based deepfake detectors, namely XceptionNet~\cite{chollet2017xception} and MesoNet~\cite{afchar2018mesonet}. 
These detectors are CNN-based classification models that independently classify each frame as either real or fake.
Table \ref{tab:2dframes} presents the performance of both models against state-of-the-art white-box adversarial attacks. 
For robust black-box (RBB) attacks, we generated perturbations from pre-trained clean models without accessing their parameters. 
This allows us to evaluate the robustness of the detectors in scenarios where they are not aware of each other internal architecture or parameters.

\begin{figure*}[t]
  \centering
  
  \includegraphics[scale=0.28]{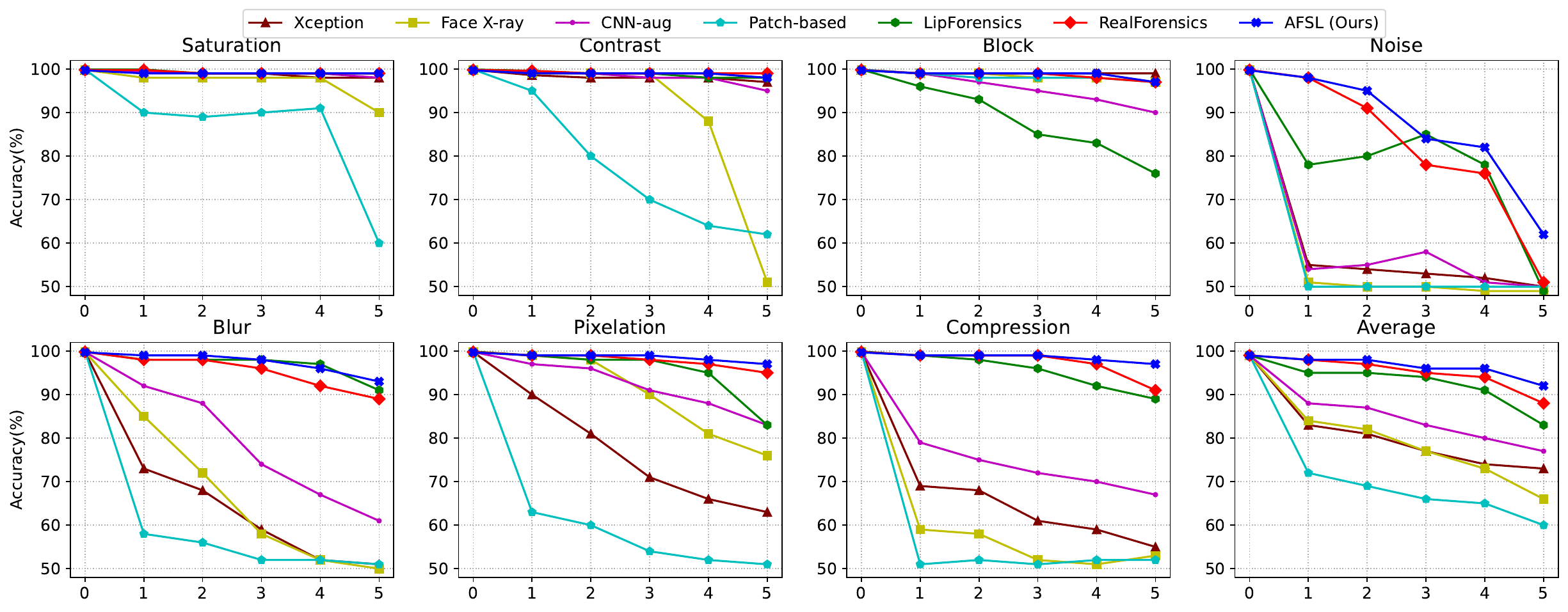}
  \caption{Robustness to unseen distortions: Video level AUC scores (\%) varying with the severity level of different distortions. Average is the mean value at each severity level. }
  \label{fig:plots}
\end{figure*}
\subsection{Robustness to Common Distortions}
In addition to robust generalization across different manipulations and resistance against adversarial attacks, deepfake detectors must also withstand common distortions that videos may encounter online.
To assess the robustness of the detector against unfamiliar distortions, we follow the settings of~\cite{haliassos2022RealForensics,haliassos2021lips}.
During training on the FF++ dataset, we limit the augmentation techniques to horizontal flipping and random cropping for grayscale inputs. This approach helps prevent any potential interactions between the training distortions and those used during testing.
Following~\cite{jiang2020deeperforensics}, we use seven different types of distortions with five levels of severity.
Figure \ref{fig:plots} presents the results of each distortion with five levels of severity using the proposed method and state-of-the-art methods.
The proposed method outperforms both frame-based and sequence-based methods. 
The inclusion of an adversarial training term in the loss function acts as regularization to increase model robustness against common distortions, which is in line with the findings of Kireev et al.~\cite{kireev2022effectiveness}.
In comparison to previous methods, the proposed approach performs well under most conditions, highlighting its effectiveness in tackling distortions commonly encountered in real-world scenarios.
\section{Ablation Study}
In this section, we analyze various components of our proposed method to comprehend the factors contributing to its performance. 
We conduct ablation experiments under PGD adversarial attack to inspect its robust generalization. 
The training is performed on FaceForensics++, and we evaluate the model on FaceShifter and DeeperForensics datasets, reporting the AUC score. 
Table \ref{tab:ablation} displays the results of the ablation study.
We use deepfake classification loss as the first term in the loss function as S1 to train the model without any adversarial training. 
Unfortunately, this model proves to be inadequate in defending against adversarial attacks.
Next, by training the model with the S2 setting without the similarity regularization, we can significantly improve the AUC scores against adversarial attacks. Finally, with all three loss components in the S3 setting, we can further improve the robustness by about 2\% in AUC score.
\setlength{\tabcolsep}{8.pt}
\begin{table}[t]
    \centering
    \caption{Impact of various components. Robust AUC (\%) on FaceShifter (FSh) and DeeperForensics (DFo) using PGD10 adversarial attack.}
    \begin{tabular}{lccccc}
    \toprule
         Settings&\multicolumn{3}{c}{\textbf{Losses}}&\multicolumn{2}{c}{\textbf{AUC (\%)}}\\
         \cmidrule(lr){2-4}\cmidrule(lr){5-6}
         &$\mathcal{L}_{\text{dcl}}$&$\mathcal{L}_{\text{asl}}$& $\mathcal{L}_{\text{srl}}$&DFo&FSh  \\
         \midrule
         S1&\cmark&\xmark&\xmark&1.0&1.0\\
         S2 &\cmark&\cmark&\xmark & 81.3 &79.1\\
         S3&\cmark &\cmark&\cmark& \textbf{83.6}&\textbf{81.7}\\

         \bottomrule
    \end{tabular}
    
    \label{tab:ablation}
\end{table}
\section{Conclusion and Future work}
This paper introduces a novel approach Adversarial Feature Similarity Learning (AFSL) for enhancing the robustness of a deepfake detector against adversarial attacks. 
We propose an adversarially robust loss function, specifically designed to detect fake videos even when subjected to deliberate adversarial perturbations.
Our experimental results demonstrate the effectiveness of the proposed method under unperturbed input but also against common distortions.
\hl{The future work will consider self-supervised learning using the proposed loss function, such as pairing real and fake samples in self-supervised adversarial defense. }
\section{ACKNOWLEDGMENT}
This research is supported by National Science and Technology Council, Taiwan (R.O.C), under the grant number of NSTC-111-2634-F-002-022, 110-2221-E-001-009-MY2, 112-2634-F-001-001-MBK, and Academia Sinica under the grant number of AS-CDA-112-M09. In addition, we would like to express our gratitude for the valuable contributions and guidance from these organizations, which have been instrumental in achieving the goals of this research.

\bibliographystyle{splncs04}
\bibliography{reference}

\begin{thebibliography}{10}
\providecommand{\url}[1]{\texttt{#1}}
\providecommand{\urlprefix}{URL }
\providecommand{\doi}[1]{https://doi.org/#1}

\bibitem{afchar2018mesonet}
Afchar, D., Nozick, V., Yamagishi, J., Echizen, I.: Mesonet: a compact facial
  video forgery detection network. In: WIFS. pp.~1--7 (2018)

\bibitem{alnaim2023dffmd}
Alnaim, N.M., Almutairi, Z.M., Alsuwat, M.S., Alalawi, H.H., Alshobaili, A.,
  Alenezi, F.S.: Dffmd: A deepfake face mask dataset for infectious disease era
  with deepfake detection algorithms. IEEE Access pp. 16711--16722 (2023)

\bibitem{carlini2017adversarial}
Carlini, N., Wagner, D.: Adversarial examples are not easily detected:
  Bypassing ten detection methods. In: AIS. pp. 3--14 (2017)

\bibitem{chai2020makes}
Chai, L., Bau, D., Lim, S.N., Isola, P.: What makes fake images detectable?
  understanding properties that generalize. In: ECCV. pp. 103--120 (2020)

\bibitem{chen2022towards}
Chen, G., Zhao, Z., Song, F., Chen, S., Fan, L., Wang, F., Wang, J.: Towards
  understanding and mitigating audio adversarial examples for speaker
  recognition. TDSC  (2022)

\bibitem{chollet2017xception}
Chollet, F.: Xception: Deep learning with depthwise separable convolutions. In:
  CVPR. pp. 1251--1258 (2017)

\bibitem{Deepfakes}
Deepfakes: faceswap. In: GitHub. ( Accessed: 14.06.2023) (2017),
  \url{https://github.com/deepfakes/faceswap}

\bibitem{Dong_2023_CVPR}
Dong, S., Wang, J., Ji, R., Liang, J., Fan, H., Ge, Z.: Implicit identity
  leakage: The stumbling block to improving deepfake detection generalization.
  In: CVPR. pp. 3994--4004 (2023)

\bibitem{frank2020leveraging}
Frank, J., Eisenhofer, T., Sch{\"o}nherr, L., Fischer, A., Kolossa, D., Holz,
  T.: Leveraging frequency analysis for deep fake image recognition. In: ICML.
  pp. 3247--3258 (2020)

\bibitem{gandhi2020adversarial}
Gandhi, A., Jain, S.: Adversarial perturbations fool deepfake detectors. In:
  IJCNN. pp.~1--8 (2020)

\bibitem{gao2021information}
Gao, G., Huang, H., Fu, C., Li, Z., He, R.: Information bottleneck
  disentanglement for identity swapping. In: CVPR. pp. 3404--3413 (2021)

\bibitem{gao2021high}
Gao, Y., Wei, F., Bao, J., Gu, S., Chen, D., Wen, F., Lian, Z.: High-fidelity
  and arbitrary face editing. In: CVPR. pp. 16115--16124 (2021)

\bibitem{goodfellow2014explaining}
Goodfellow, I.J., Shlens, J., Szegedy, C.: Explaining and harnessing
  adversarial examples. ICLR  (2015)

\bibitem{guan2022delving}
Guan, J., Zhou, H., Hong, Z., Ding, E., Wang, J., Quan, C., Zhao, Y.: Delving
  into sequential patches for deepfake detection. arXiv preprint
  arXiv:2207.02803  (2022)

\bibitem{haliassos2022RealForensics}
Haliassos, A., Mira, R., Petridis, S., Pantic, M.: Leveraging real talking
  faces via self-supervision for robust forgery detection. In: CVPR. pp.
  14950--14962 (2022)

\bibitem{haliassos2021lips}
Haliassos, A., Vougioukas, K., Petridis, S., Pantic, M.: Lips don't lie: A
  generalisable and robust approach to face forgery detection. In: CVPR. pp.
  5039--5049 (2021)

\bibitem{hou2023evading}
Hou, Y., Guo, Q., Huang, Y., Xie, X., Ma, L., Zhao, J.: Evading deepfake
  detectors via adversarial statistical consistency. In: CVPR. pp. 12271--12280
  (2023)

\bibitem{hussain2021adversarial}
Hussain, S., Neekhara, P., Jere, M., Koushanfar, F., McAuley, J.: Adversarial
  deepfakes: Evaluating vulnerability of deepfake detectors to adversarial
  examples. In: WACV. pp. 3348--3357 (2021)

\bibitem{jiang2020deeperforensics}
Jiang, L., Li, R., Wu, W., Qian, C., Loy, C.C.: Deeperforensics-1.0: A
  large-scale dataset for real-world face forgery detection. In: CVPR. pp.
  2889--2898 (2020)

\bibitem{ACL2020}
Jiang, Z., Chen, T., Chen, T., Wang, Z.: Robust pre-training by adversarial
  contrastive learning. NIPS pp. 16199--16210 (2020)

\bibitem{karras2019style}
Karras, T., Laine, S., Aila, T.: A style-based generator architecture for
  generative adversarial networks. In: CVPR. pp. 4401--4410 (2019)

\bibitem{kawa2023defense}
Kawa, P., Plata, M., Syga, P.: Defense against adversarial attacks on audio
  deepfake detection. In: Interspeech (2023)

\bibitem{khan2017text}
Khan, S., Thainimit, S., Kumazawa, I., Marukatat, S.: Text detection and
  recognition on traffic panel in roadside imagery. In: 2017 8th International
  Conference of Information and Communication Technology for Embedded Systems
  (IC-ICTES). pp.~1--6. IEEE (2017)

\bibitem{kireev2022effectiveness}
Kireev, K., Andriushchenko, M., Flammarion, N.: On the effectiveness of
  adversarial training against common corruptions. In: UAI. pp. 1012--1021
  (2022)

\bibitem{li2020advancing}
Li, L., Bao, J., Yang, H., Chen, D., Wen, F.: Advancing high fidelity identity
  swapping for forgery detection. In: CVPR. pp. 5074--5083 (2020)

\bibitem{Li_2023_CVPR}
Li, Z., Yin, B., Yao, T., Guo, J., Ding, S., Chen, S., Liu, C.: Sibling-attack:
  Rethinking transferable adversarial attacks against face recognition. In:
  CVPR. pp. 24626--24637 (2023)

\bibitem{Liang_2023_CVPR}
Liang, K., Xiao, B.: Styless: Boosting the transferability of adversarial
  examples. In: CVPR. pp. 8163--8172 (2023)

\bibitem{Liu_2023_WACV}
Liu, B., Liu, B., Ding, M., Zhu, T., Yu, X.: Ti2net: Temporal identity
  inconsistency network for deepfake detection. In: WACV. pp. 4691--4700 (2023)

\bibitem{lo2021defending}
Lo, S.Y., Patel, V.M.: Defending against multiple and unforeseen adversarial
  videos. TIP pp. 962--973 (2021)

\bibitem{madry2017towards}
Madry, A., Makelov, A., Schmidt, L., Tsipras, D., Vladu, A.: Towards deep
  learning models resistant to adversarial attacks. ICLR  (2018)

\bibitem{menon2021logitadjustment}
Menon, A.K., Jayasumana, S., Rawat, A.S., Jain, H., Veit, A., Kumar, S.:
  Long-tail learning via logit adjustment. ICLR  (2021)

\bibitem{mumcu2022adversarial}
Mumcu, F., Doshi, K., Yilmaz, Y.: Adversarial machine learning attacks against
  video anomaly detection systems. In: CVPR. pp. 206--213 (2022)

\bibitem{advfake}
Neekhara, P.: Adversarialdeepfake. In: GitHub. ( Accessed: 14.06.2023) (2019),
  \url{https://github.com/paarthneekhara/AdversarialDeepFakes}

\bibitem{neekhara2021adversarial}
Neekhara, P., Dolhansky, B., Bitton, J., Ferrer, C.C.: Adversarial threats to
  deepfake detection: A practical perspective. In: CVPR. pp. 923--932 (2021)

\bibitem{neekhara2019universal}
Neekhara, P., Hussain, S., Pandey, P., Dubnov, S., McAuley, J., Koushanfar, F.:
  Universal adversarial perturbations for speech recognition systems. arXiv
  preprint arXiv:1905.03828  (2019)

\bibitem{qin2019imperceptible}
Qin, Y., Carlini, N., Cottrell, G., Goodfellow, I., Raffel, C.: Imperceptible,
  robust, and targeted adversarial examples for automatic speech recognition.
  In: ICML. pp. 5231--5240 (2019)

\bibitem{rossler2019faceforensics++}
Rossler, A., Cozzolino, D., Verdoliva, L., Riess, C., Thies, J., Nie{\ss}ner,
  M.: Faceforensics++: Learning to detect manipulated facial images. In: CVPR.
  pp. 1--11 (2019)

\bibitem{shahzad2022lip}
Shahzad, S.A., Hashmi, A., Khan, S., Peng, Y.T., Tsao, Y., Wang, H.M.: Lip sync
  matters: A novel multimodal forgery detector. In: APSIPA. pp. 1885--1892
  (2022)

\bibitem{songja2023deepfake}
Songja, R., Promboot, I., Haetanurak, B., Kerdvibulvech, C.: Deepfake ai
  images: should deepfakes be banned in thailand? AI and Ethics pp. 1--13
  (2023)

\bibitem{spivak2018deepfakes}
Spivak, R.: " deepfakes": The newest way to commit one of the oldest crimes.
  HeinOnline p.~339 (2018)

\bibitem{tran2019video}
Tran, D., Wang, H., Torresani, L., Feiszli, M.: Video classification with
  channel-separated convolutional networks. In: ICCV. pp. 5552--5561 (2019)

\bibitem{tran2018closer}
Tran, D., Wang, H., Torresani, L., Ray, J., LeCun, Y., Paluri, M.: A closer
  look at spatiotemporal convolutions for action recognition. In: CVPR. pp.
  6450--6459 (2018)

\bibitem{wang2021understanding}
Wang, H., He, F., Peng, Z., Shao, T., Yang, Y.L., Zhou, K., Hogg, D.:
  Understanding the robustness of skeleton-based action recognition under
  adversarial attack. In: CVPR. pp. 14656--14665 (2021)

\bibitem{yadlin2021whose}
Yadlin-Segal, A., Oppenheim, Y.: Whose dystopia is it anyway? deepfakes and
  social media regulation. Convergence pp. 36--51 (2021)

\bibitem{yang2021defending}
Yang, C., Ding, L., Chen, Y., Li, H.: Defending against gan-based deepfake
  attacks via transformation-aware adversarial faces. In: IJCNN. pp.~1--8
  (2021)

\bibitem{zhang2019theoretically}
Zhang, H., Yu, Y., Jiao, J., Xing, E., El~Ghaoui, L., Jordan, M.: Theoretically
  principled trade-off between robustness and accuracy. ICML pp. 7472--7482
  (2019)

\bibitem{zheng2021exploring}
Zheng, Y., Bao, J., Chen, D., Zeng, M., Wen, F.: Exploring temporal coherence
  for more general video face forgery detection. In: ICCV. pp. 15044--15054
  (2021)

\end{thebibliography}
\end{document}